\documentclass[sigconf]{acmart}%{sig-alternate} 
\usepackage{color}
\usepackage{url}
\usepackage{verbatim} % allows multiline comments
\usepackage{subfigure} % for the subfloat in the figs
\usepackage{multirow}
\usepackage{algorithm}
\usepackage[noend]{algorithmic}
\usepackage{xspace}
\usepackage{graphicx}
\usepackage{epsfig}
\usepackage{amsmath}
\usepackage{amssymb}
\usepackage{enumerate}
\usepackage{enumitem}

\usepackage{dcolumn}

\usepackage{graphicx,psfrag,amsmath,amsfonts,appendix}
\usepackage{algorithm}

\newcommand{\reals}{{\mbox{\bf R}}}

\newcommand{\hide}[1]{}
\newcommand{\xhdr}[1]{\vspace{1.7mm}\noindent{{\bf #1.}}}

\newcommand{\eg}{\emph{e.g.}}
\newcommand{\ie}{\emph{i.e.}}

\newcommand{\symmpd}[1]{{\mathbf{S}}^{#1}_{++}}  % p-dimensional symmetric positive matrices 
\newcommand{\norm}[1]{\left\|#1\right\|}
\newcommand{\twonorm}[1]{\norm{#1}_2}

\newcommand{\odnorm}[1]{\norm{#1}_{\mathrm{od},1}}
\newcommand{\fronorm}[1]{\norm{#1}_{F}}

\newcommand{\minimize}[1]{\underset{#1}{\mathop{\rm minimize}} \qquad}

\newcommand{\Tr}[1]{\mathop{\mathbf{Tr}(#1)}}
\newcommand{\subjectto}{\mathop{\rm subject \ to \qquad}}
\newcommand{\bmat}[1]{\begin{bmatrix}#1\end{bmatrix}}
\newcommand{\argmin}[1]{\underset{#1}{\mathop{\rm argmin ~}}}

\newcommand{\prox}{\mathbf{prox}}

\graphicspath{{./FIG/}}

\newfont{\mycrnotice}{ptmr8t at 7pt}
\newfont{\myconfname}{ptmri8t at 7pt}

\begin{document}

\copyrightyear{2017} 
\acmYear{2017} 
\setcopyright{acmlicensed}
\acmConference{KDD '17}{August 13-17, 2017}{Halifax, NS, Canada}\acmPrice{15.00}\acmDOI{10.1145/3097983.3098037}
\acmISBN{978-1-4503-4887-4/17/08}

\fancyhead{}
\settopmatter{printacmref=false, printfolios=false}

\title{Network Inference via the Time-Varying Graphical Lasso}

\author{David Hallac, Youngsuk Park, Stephen Boyd, Jure Leskovec}
\affiliation{%
  \institution{Stanford University}
}
\email{{hallac, youngsuk, boyd, jure}@stanford.edu}

\begin{abstract}
% !TEX root = paper-netinf.tex
Many important problems can be modeled as a system of interconnected entities, where each entity is recording time-dependent observations or measurements. In order to spot trends, detect anomalies, and interpret the temporal dynamics of such data, it is essential to understand the relationships between the different entities and how these relationships evolve over time. In this paper, we introduce the \emph{time-varying graphical lasso (TVGL)}, a method of inferring time-varying networks from raw time series data. We cast the problem in terms of estimating a sparse time-varying inverse covariance matrix, which reveals a dynamic network of interdependencies between the entities. Since dynamic network inference is a computationally expensive task, we derive a scalable message-passing algorithm based on the Alternating Direction Method of Multipliers (ADMM) to solve this problem in an efficient way. We also discuss several extensions, including a streaming algorithm to update the model and incorporate new observations in real time. Finally, we evaluate our TVGL algorithm on both real and synthetic datasets, obtaining interpretable results and outperforming state-of-the-art baselines in terms of both accuracy and scalability.

\end{abstract}

\maketitle

% \vspace{1mm}
%  \noindent {\bf Categories and Subject Descriptors:} H.2.8 {\bf
% [Database Management]}: Database applications---{\it Data mining}

% \noindent {\bf General Terms:} Algorithms; Experimentation.

% %\noindent {\bf Keywords:} Network Inference, Time Series Analysis, Convex Optimization, ADMM, Graphical Lasso.
% \noindent {\bf Keywords:} Network Inference, Time Series Analysis, ADMM.

\section{Introduction}
\label{sec:intro}
% !TEX root = paper-netinf.tex

\begin{figure}[t]
    \centering
    \includegraphics[width=0.9\linewidth]{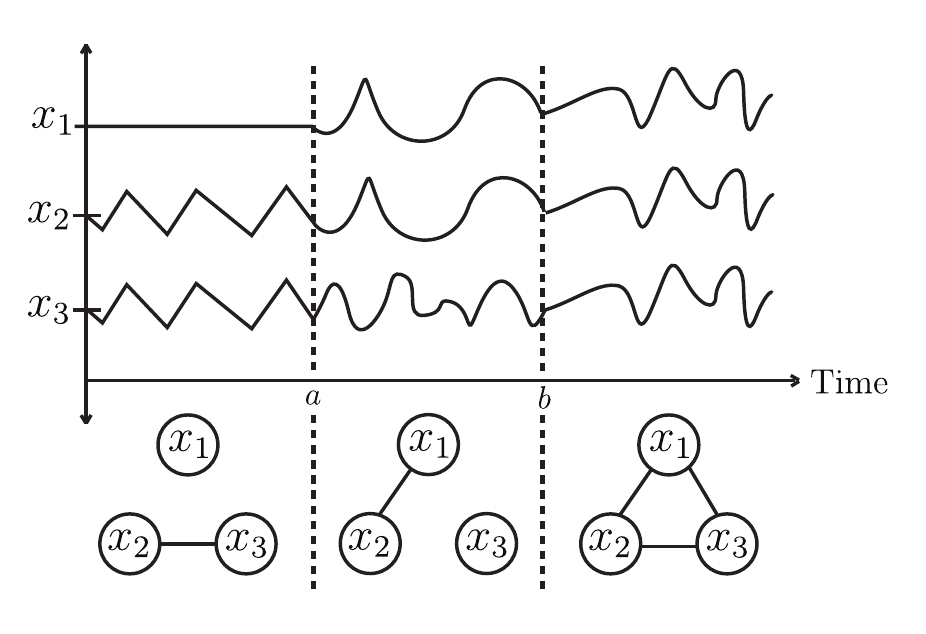}
    \vspace{-5mm}
    \caption{Three sensors with associated time series readings. Based on such data, 
    we infer a time-varying network that reveals 1) the dependencies between the different sensors, and 2) when and how these dependencies change over time.}
    \vspace{-5mm}
    \label{fig:toyExample}
\end{figure}

Applications in many settings, ranging from neurological connectivity patterns~\cite{MHSLAM:14} to financial markets~\cite{N:11} and social network analysis~\cite{AX:09,myers10connie}, contain massive sequences of multivariate timestamped observations. Such data can often be modeled as a network of interacting entities, where each entity is a node associated with a time series of data points. 
In these dependency networks, also known as Markov random fields (MRFs) \cite{KF:09, RH:05, WK:13}, an edge represents a partial correlation, or a direct effect (holding all other nodes constant) between two entities. 
An important problem that arises in many applications is using observational data to infer these relationships (\ie, edges) and their evolution over time. 
In particular, it is necessary to understand how the structure of these complex systems changes over a period of interest (Figure~\ref{fig:toyExample}). 
For example, in financial markets, companies can be represented as nodes, and each acts like a ``sensor'' 
recording a time series of its stock price. 
By understanding the relationships within the network and their evolution over time, one can detect anomalies, spot trends, classify events,  
forecast future behavior, and solve many other problems at the intersection of time series analysis and network science.

To learn these networks, one can model the relationships between the entities through an underlying inverse covariance matrix that changes over time. 
Doing so allows for inference of a dynamic undirected network, with nodes representing the different entities and edges defining the coupling between them. More precisely, given a multivariate sequence of readings, one can estimate the true inverse covariance matrix $\Sigma^{-1}(t)$ (which changes over time), assuming a Gaussian distribution. The focus is specifically on the inverse covariance because of its increased interpretability: if $\Sigma^{-1}_{ij}(t) = 0$ then, given the values of all the other entities (\ie, nodes), $i$ and $j$ are conditionally independent at time $t$~\cite{L:96}. Therefore, the inferred network has an edge between $i$ and $j$ at time $t$ if $\Sigma^{-1}_{ij}(t) \neq 0$, denoting a structural dependency between these two entities at that moment in time. In the static case, where $\Sigma^{-1}$ is constant, this inference is known as the graphical lasso problem~\cite{FHT:08,YL:07}. While many efficient algorithms exist for solving the graphical lasso~\cite{BED:08, hsieh2013big}, such methods do not generalize to the time-varying case.

Inferring dynamic networks is challenging mainly because it is difficult to simultaneously estimate both the network itself and the change in its structure over time. This is in part due to the fact that networks can exhibit many different types of changes. 
The range of possibilities includes a sudden shift of the entire network structure, a single node rewiring all of its connections, or even just one or two edges changing in the whole network.
Therefore, any method must be general enough to discover many types of evolutionary patterns, while also being powerful enough to learn this temporal structure over very long time series.
As such, solving for time-varying networks is computationally expensive, especially compared to time-invariant inference~\cite{DWW:14, MLFWL:14}. 
There are more parameters, additional coupling, and more complicated dynamics. 
Standard methods have trouble scaling to large examples, so novel algorithms and techniques are required.

\xhdr{Present Work}
In this paper, we formulate time-varying network inference as a convex optimization problem. 
Thus, our two primary contributions are in formally defining this problem, which we call the \emph{time-varying graphical lasso} (TVGL), and in deriving a scalable optimization method to solve it.
Given a sequence of multivariate observations, the TVGL estimates the inverse covariance $\Sigma^{-1}(t)$ of the data to simultaneously achieve three goals: (1) matching the empirical observations, (2) sparsity, and (3) temporal consistency. Matching the empirical observations ensures that our estimated network is well-supported by the data. A sparse inverse covariance prevents overfitting and provides interpretable results via a sparse graphical representation \cite{BED:08,FHT:08}.  
Temporal consistency is the idea that, most of the time, adjacent timestamps should have very similar (or even identical) estimations of the network.
We impose a temporal penalty to limit how the network can evolve, where different penalties induce different dynamics. We suggest five specific penalties that allow us to model different types of temporal variation (or evolutionary patterns) in networks: e.g., smoothly varying networks, rare-but-large-scale shifts, or a single node rewiring its connections.
Then, since no scalable methods exist for solving our TVGL problem, we develop a message-passing algorithm using the alternating direction method of multipliers (ADMM)~\cite{BPCPE:11}.
We derive closed-form solutions for the ADMM subproblems, including one for each of the five unique penalty types, to further speed up the runtime. We also discuss several extensions, 
including one to convert our approach into 
a streaming algorithm that can quickly incorporate new data and update our estimate in real time.

We then apply our TVGL method to both real and synthetic datasets. First, we test accuracy and scalability on synthetic examples with known ground truth networks. Our TVGL method leads to accuracy improvements of up to 92\% over two state-of-the-art baselines. Furthermore, our ADMM-based implementation is several orders of magnitude faster than other solution methods, able to solve for 5 million unknown variables in under 12 minutes (while other solvers take several hours for even 50 thousand unknowns). Finally, we analyze two real-world case studies, using financial and automobile sensor data, to demonstrate how the TVGL approach can find meaningful insights, understandable structure, and different types of evolutionary patterns in time series data.

\xhdr{Related Work}
This work relates to recent advancements in both graphical models and convex optimization. Inferring static networks via the graphical lasso is a well-studied topic~\cite{BED:08,DWW:14,FHT:08,YL:07}. However, previous work on dynamic inference has only focused on a kernel method~\cite{zhou2010time} or an $\ell_1$-fused penalty~\cite{kolar2010estimating, MHSLAM:14, WA:15}. One of the main contributions of our approach is that it is able to model many different types of network evolutionary patterns, for example a small set of edges rewiring, a single node changing all its edges, or the entire network restructuring at a single time step. This opens up a variety of new applications, several of which we examine in Sections 6 and 7. Whereas previous work allows for only one time-varying pattern~\cite{kolar2010estimating, MHSLAM:14, WA:15}, we show how the selection of the proper evolutionary penalty is a very important parameter for obtaining accurate results.  
We also propose several extensions, including a streaming approach, that to the best of our knowledge have not been explored in the literature.

To solve our TVGL problem, we develop an ADMM-based algorithm~\cite{BPCPE:11} so that the same framework can incorporate all of the different penalty types that we study. This is necessary because, even though many problem-specific methods exist to solve the standard (static) graphical lasso~\cite{FHT:08, hsieh2013big}, no ready-to-use methods exist for the time-varying case. 
To make our algorithm more scalable, we rewrite the ADMM subproblems in terms of proximal operators \cite{DWW:14, MHSLAM:14,SMG:10}. This allows us to take advantage of known properties and solution methods to 
derive closed-form ADMM updates~\cite{BV:04,PB:14}, which speed up our solver by several orders of magnitude over a naive ADMM implementation (Section 6.2).

Another common method of time series analysis is the Kalman filter \cite{grewal2011kalman}, a special type of dynamic factor model \cite{molenaar1985dynamic}. Although used in similar domains, these models are fundamentally different from our network inference approach.
They are typically used to predict values of unknown variables, given all previous data. Our algorithm instead learns the underlying dynamic graphical structure of the variables \cite{KF:09, L:96}. Thus, our method helps remove the effects of noisy data and adds interpretability to the results. 
Additionally, we note the difference between our work here and information cascade-based network inference~\cite{gomez12netinf,myers10connie}, which assumes a viral spreading process over the nodes and aims to infer the links based on the node infection times. We instead aim to uncover the dependency structure (\ie, inverse covariance matrix) based on multivariate time series observations.

\begin{figure*}
    \centering
    \includegraphics[width=0.9\linewidth]{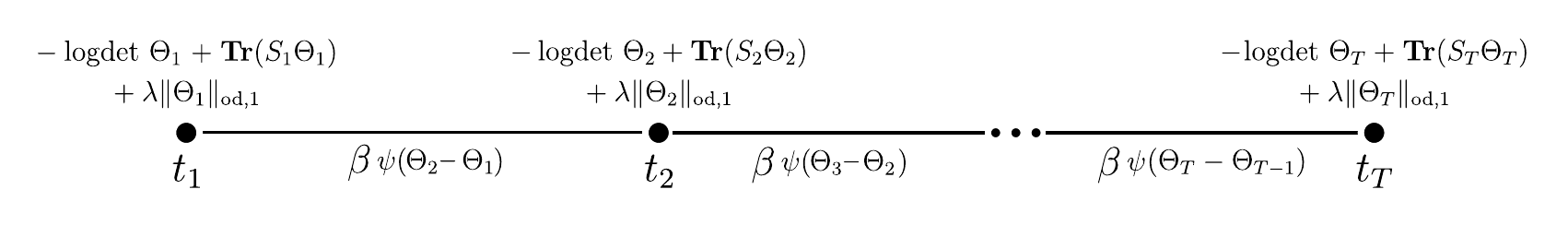}
    \vspace{-6mm}
    \caption{Dynamic network inference can be thought of as an optimization problem on a chain graph, where each node objective solves for a \emph{network slice} at each timestamp, and edge objectives define the penalties that enforce temporal consistency.}
    \vspace{-3mm}
    \label{fig:chainGraph}
\end{figure*}

\section{Problem Definition}
\label{sec:original}
% !TEX root = paper-netinf.tex

Consider a sequence of multivariate observations in $\reals^p$ sampled from a distribution $x \thicksim N(0,\Sigma(t))$. Observations come in at times $0 \leq t_1 \leq \cdots \leq t_T$, where at each time $t_i$, there are $n_i \geq 1$ different observation vectors over the readings of all the nodes. (For now we assume that the readings are synchronous, where $t_i - t_{i-1}$ is a constant value for all $i$, but in Section 3.1 we extend this approach to asynchronous observations.) With these samples, we aim to estimate the underlying covariance matrix $\Sigma(t)$, which can change over time. That is, given a changing underlying distribution, we attempt to estimate it based on a sequence of observations.
Here, we formulate a convex optimization problem to infer a sequence of sparse inverse covariance matrices, $\Theta_i = \Sigma(t_i)^{-1}$, one for each $t_i$. These estimates are based on local empirical observation vector(s), as well as coupling constraints with neighboring timestamps' covariance estimates.
Recall that a sparse inverse covariance allows us to encode conditional independence between different variables~\cite{KF:09}.

\xhdr{Inferring Static Networks}
We first consider static inference, which is equivalent to the graphical lasso problem \cite{BED:08,FHT:08,YL:07}, and then build on it to extend to dynamic networks. 
In the static case, $\Sigma(t)$ is constant for all $t$.  Given a series of multivariate readings, this can be written as
\begin{align}
	\small
	\mathrm{minimize} & - l(\Theta) + \lambda \odnorm{\Theta},
	\label{graphLasso}
\end{align}
where $\odnorm{\Theta}$ is a seminorm of $\Theta$, the off-diagonal $\ell_1$-norm, $\sum_{i\neq j} |\Theta_{ij}|$. 
This lasso penalty enforces element-wise sparsity in our solution for $\Theta$, regulated by the trade-off parameter $\lambda \geq 0$. 
In Problem \eqref{graphLasso}, $l(\Theta)$ is the log likelihood of $\Theta$ (up to a constant and scale) \cite{WJ:06,YL:07},
\begin{align*}
	\small
	l(\Theta) = n(\log\det \Theta - \Tr{S \Theta}),
\end{align*}
where $\Theta$ must be symmetric positive-definite ($\symmpd{p}$), $S$ is the empirical covariance $\frac{1}{n} \sum_{i=1}^{n} x_i x_i^T$, $n$ is the number of observations, and $x_i$ are the different samples. When $S$ is invertible, $l(\Theta)$ encourages $\Theta$ to be close to $S^{-1}$.

\xhdr{Inferring Dynamic Networks}
In order to infer a time-varying sequence of networks, we extend the above approach and allow $\Sigma(t)$ to vary over time.
To do so we set up a sequence of graphical lasso problems, each coupled together in a chain to penalize deviations in the estimations, which we call the \emph{time-varying graphical lasso} (TVGL). We solve for $\Theta = (\Theta_1,\ldots,\Theta_T)$, our estimated inverse covariance matrices at times $t_1, \ldots, t_T$,
\begin{align}
\small
\minimize{\Theta \in \symmpd{p}} &\sum_{i=1}^T - l_i(\Theta_i) + \lambda \odnorm{\Theta_i} + \beta \sum_{i=2}^T \psi(\Theta_{i}-\Theta_{i-1}).
\label{original}
\end{align}
Here, $l_i(\Theta_i) = n_i(\log\det \Theta_i - \Tr{S_i \Theta_i})$, $\beta \geq 0$, and $\psi(\Theta_{i} - \Theta_{i-1})$ is a convex penalty function, minimized at $\psi(0)$, which encourages similarity between $\Theta_{t-1}$ and $\Theta_t$. We examine in Section 2.1 how different $\psi$'s enforce different behaviors. Note that Problem \eqref{original} can be viewed as an optimization problem defined on a chain graph, where each element solves for a different $\Theta_i$, 
as shown in Figure 2.

\xhdr{Empirical Covariance}
The log-likelihood $l_i$ depends on $S_i$, the empirical covariance at time $t_i$. In high-dimensional settings, where the number of dimensions $p$ is larger than the number of observations $n_i$, $S_i$ will be rank deficient and therefore non-invertible. However, our method is well-suited to overcome this problem of an insufficient number of observations. By enforcing structural similarity, each $\Theta_i$ borrows strength from the fact that neighboring network estimates should be similar, or even identical, across time. In the extreme case, we are able to estimate a network at a time where there is only one observation. The empirical covariance, $x_i x_i^T$, is rank 1, but between the sparsity and the temporal consistency penalties, our approach can still infer an accurate estimate of the network at that snapshot in time.

\xhdr{Regularization Parameters $\lambda$ and $\beta$}
Problem \eqref{original} has two parameters, $\lambda$ and $\beta$, which define important values on two trade-off curves. $\lambda$ determines the sparsity level of the network: small values will better match the empirical data, but will lead to very dense networks, which are less interpretable and often overfit. 
$\beta$ determines how strongly correlated neighboring covariance estimations should be. A small $\beta$ will lead to $\Theta$'s which fluctuate from estimate-to-estimate, whereas large $\beta$'s lead to smoother estimates over time. As $\beta \rightarrow \infty$, the temporal deviation penalty gets so large that Problem \eqref{original} turns into the original graphical lasso, since a constant $\Theta$ will be the solution across the entire time series.

\subsection{Encoding Network Evolutionary Patterns}
Different types of penalty functions $\psi$ allow us to enforce different behaviors in the evolution of the network structure. In particular, if we have an expectation about how the underlying network may change over time, we are able to encode it into $\psi$. Here, we define several common temporal patterns and their associated penalty functions. In Sections 6 and 7, we use these to analyze the time-varying dynamics of multiple real and synthetic datasets. For these penalties, note that $\psi$ can sometimes be split into a sum of column-norms. As such, we refer to the $j$-th column of a matrix $X$ as $[X]_j$.
\begin{itemize}[nolistsep, leftmargin=.1in]
    \item \textbf{A few edges changing at a time ---} Setting $\psi(X) = \sum_{i,j} |X_{i,j}|$, an element-wise $\ell_1$ penalty, encourages neighboring graphs to be identical \cite{DWW:14,YLSWY:15}. When the $ij$-th edge of the network ``breaks'', or is different at two neighboring times, this penalty still encourages the rest of the graph to remain \emph{exactly} the same. As a result, this penalty is best used in cases where we expect only a handful of edges --- at most --- to change at a time.
    \item \textbf{Global restructuring ---} Setting $\psi(X) = \sum_{j} \|[X]_j\|_2$, a group lasso $\ell_2$ penalty, causes the entire graph to restructure at a select few timestamps, while at all other times, the graph remains piecewise constant \cite{DWW:14,HLB:15}. This is useful for event detection and time-series segmentation, as it finds exact times where there is a regime-change, or shift, in the underlying covariance matrix.
    \item \textbf{Smoothly varying over time ---} Setting $\psi(X) = \sum_{i,j} X_{i,j}^2$, a Laplacian penalty, causes smooth transitions of the graphical model from timestamp to timestamp. Adjacent graphs will often differ by small amounts, but severe deviations are largely penalized so they rarely occur \cite{WSZS:06}. This penalty is best used when we want to come up with a smoothly varying estimate over time.
    \item \textbf{Block-wise restructuring ---} Setting $\psi(X)$ to an $\ell_{\infty}$ penalty, $\psi(X) = \sum_{j} \left( \max_{i} |X_{i,j}| \right)$, implies that, when an element in the inverse covariance matrix changes by $\epsilon$ at a given time, other elements are free to change by up to that same amount with no additional penalty (except for the original $\ell_{\mathrm{od},1}$ sparsity condition). This is best used when a cluster of nodes suddenly changes its internal edge structure, the rest of the network does not change.
    \item \textbf{Perturbed node ---} Setting $\psi$ to the row-column overlap penalty \cite{MCHWLF:12}, defined as $\psi(X) = \min\limits_{V: V + V^T = X} \sum\limits_{j} \|[V]_j\|_2$, yields an interesting behavior. When node $i$ has a reweighting of one edge at time $t$, this says that the node can rewire all its edges at that same time with a minimal penalty. However the rest of the graph is strongly encouraged to remain the exact same. It is best used in situations where we are looking for single nodes re-locating themselves to a new set of neighbors within the network.            
\end{itemize}

\section{Extensions}
\label{sec:streaming}
% !TEX root = paper-netinf.tex

Here, we develop three extensions of the basic time-varying graphical lasso. First, we update our approach to allow for asynchronous observations of the data. Then, we derive a method of inferring an estimate at \emph{any} time, even if there is no temporally-local data. Finally, we extend the algorithm so it can be deployed in an application with streaming data and real-time update constraints.

\xhdr{Asynchronous Observations}
Problem \eqref{original} can be extended to asynchronous settings, where samples are observed at irregularly-spaced intervals. We rewrite the problem as 
\begin{align*}
\small
\minimize{\Theta \in \symmpd{p}} \sum_{i=1}^T & -l_i(\Theta_i) + \lambda \odnorm{\Theta_i}
 + \beta \sum_{i=2}^T h_i \psi\left(\frac{\Theta_{i} - \Theta_{i-1}}{h_i}\right),
\end{align*}
where $h_i = t_i - t_{i-1}$, the interval of time between sample $i-1$ and sample $i$. 
This is similar to putting $h_i$ different intermediate $\Theta$'s between the two samples (with no loss, since there is no data). Since $\psi$ is convex, we know that the loss is minimized by evenly spreading the change in $\Theta_{i} - \Theta_{i-1}$ across these $h_i$ steps. We also know from convexity that this penalty gets smaller as $h_i$ gets larger. That is, the temporal consistency penalty is less important when consecutive samples have a large time gap between them. Of course, we can scale this penalty through the regularization parameter $\beta$ to account for the average frequency of the data.

\xhdr{Inferring Intermediate Networks}
With this approach, it is possible to infer the covariance estimation at \emph{any} time, even if there are no observations at that moment in time. This can be very useful if, for example, we want to get a granular estimate of a sharp breakpoint. To infer a network estimate at intermediate time $s$, we create a dummy node at $s$ and merge it into the chain graph by connecting it to the nearest observation in either direction, $j-1$ and $j$. To get an estimate of $\Theta_s$, we solve for
\begin{align*}
\small
\minimize{\Theta_s \in \symmpd{p}} &w(s - t_{j-1}) \psi(\Theta_{s} - \Theta_{j-1}) + w(t_{j}-s) \psi(\Theta_{j} - \Theta_{s}).
\end{align*}
For most common $\psi$'s, this problem 
has a closed-form solution. With this we can efficiently ``upsample'' and predict underlying covariances as frequently as our application requires, even if the data has a slower sampling frequency.

\xhdr{Streaming Algorithm}
In many applications, it is necessary to deploy a network inference scheme that updates in real time as new observations continue to arrive. Therefore, we require a streaming algorithm to quickly incorporate new data into our model. That is, given that we have solved for a problem with $i$ timestamps, we look for a fast way to update the solution when we receive a new observation at time $t_{i+1}$. 

One approach is to use a warm-start ADMM. With this method, we solve for the $i+1$ covariance matrices using the standard ADMM approach, but the first $i$ $\Theta$'s initialize themselves at the value of the solution to the previous problem.
However, there are no guarantees as to how many iterations warm-start ADMM may take to converge. In particular, as $i$ gets large, there is a risk that one single additional reading takes a very long time as the new information needs to propagate across the entire time series. Therefore, it may take longer to incorporate the 1000th reading than the 100th. If we want a real-time implementation of this algorithm, we need a way of guaranteeing that it takes the same amount of time to solve, regardless of the current timestamp.

\begin{figure}
    \centering
    \includegraphics[width=0.75\linewidth]{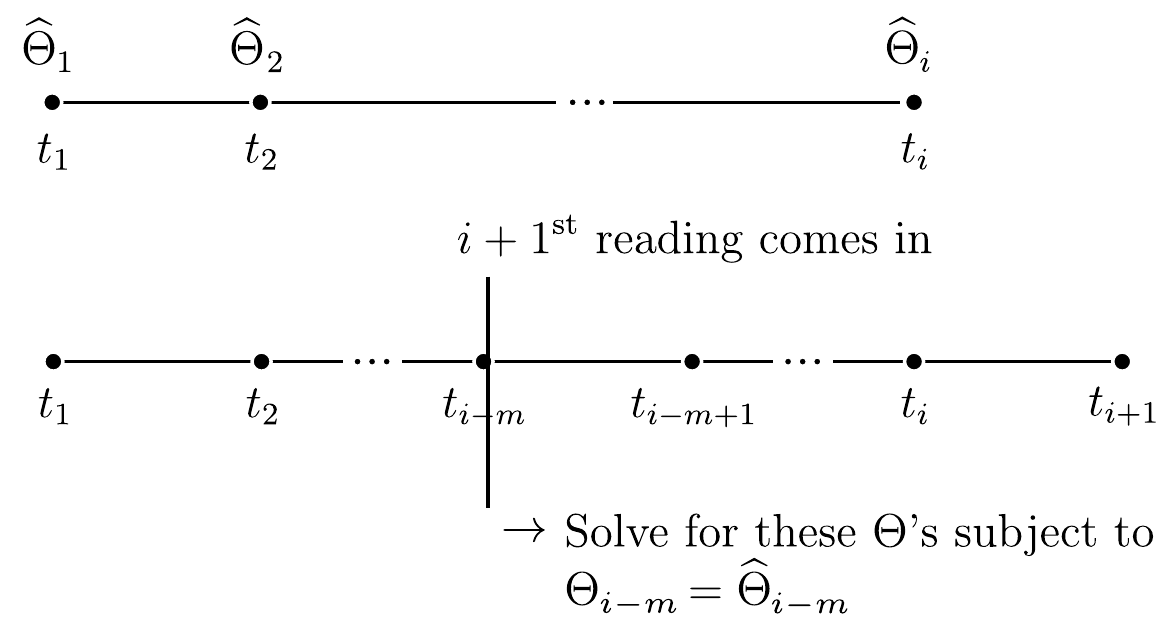}
    \vspace{-4mm}
    \caption{To quickly update our estimates when the observation at $t_{i+1}$ arrives, we re-solve for the most recent $m$ timestamps, while enforcing that $\Theta_{i-m}$ must remain the same.}
    \vspace{-4mm}
    \label{fig:streaming}
\end{figure}

We do so using a small approximation where we fix the result a certain distance in the past, and only solve for the $m$ most recent nodes (Figure \ref{fig:streaming}).
Therefore, if the $i+1$-st reading comes in, we only solve for nodes $i-m$ to $i+1$, subject to the constraint that $\Theta_{i-m}$ = $\hat\Theta_{i-m}$, where $\hat\Theta$ is the solution from when there were only $i$ timestamps. We can pick $m$ based on memory limitations, previous breakpoints, or domain expertise.

\section{Proposed Algorithm}
\label{sec:proposed}
% !TEX root = paper-netinf.tex

When inferring small networks, or those without time-varying dynamics, Problem~\eqref{original} can be solved using standard interior-point methods. However, our paper focuses on larger examples, where it is infeasible to solve the whole problem at once. Here, we propose the time-varying graphical lasso (TVGL) algorithm, based on the alternating direction method of multipliers (ADMM) \cite{BPCPE:11}, a well-established distributed convex optimization approach.
With ADMM, we split the problem up into a series of subproblems and use a message-passing algorithm to converge on the globally optimal solution. In this section, we analyze the separable subproblems and develop analytical solutions, which are fast and easy to implement, for every step in the ADMM process. To do so, we rewrite terms in the form of proximal operators \cite{PB:14}, which are defined for a matrix $A\in \reals^{m\times n}$ and the real-valued function $f(X)$ as
\begin{align}
	\small
	\prox_{\eta f}(A) = \argmin{X\in \reals^{m\times n}}\left(f(X)+1/{(2\eta)}\fronorm{X-A}^2\right).
\end{align}
The proximal operator defines a trade-off for $X$ between minimizing $f$ and being near $A$. Writing the problems in this form allows us to take advantage of well known properties to find closed-form updates for each of the ADMM subproblems. 

\subsection{ADMM Solution}
To split Problem \eqref{original} into a separable form, we introduce a consensus variable $Z=$ $\{Z_0,Z_1, Z_2\}$ = $\{(Z_{1,0},\ldots,Z_{T,0}),$ $ (Z_{1,1},\ldots,Z_{T-1,1}),$ $ (Z_{2,2},\ldots,Z_{T,2}) \}$. With this, we can rewrite Problem \eqref{original} as its equivalent problem,
\begin{align*}
\small
\minimize{} & \sum_{i=1}^T -l_i(\Theta_i) + \lambda \odnorm{Z_{i,0}} +   \beta \sum_{i=2}^T \psi(Z_{i,2} - Z_{i-1,1})\\
\subjectto &  Z_{i,0} = \Theta_i , ~ \Theta_i \in \symmpd{p} \quad \mbox{for}\quad i = 1,\ldots,T \\
 & (Z_{i-1,1}, Z_{i,2} )  = (\Theta_{i-1},\Theta_{i}) \quad \mbox{for}\quad i = 2,\ldots,T.
\end{align*}
The corresponding augmented Lagrangian \cite{H:69} becomes
\begin{align}
\small
\mathcal{L}_\rho&(\Theta, Z, U) = \sum_{i=1}^T -l(\Theta_i) + \lambda \odnorm{Z_{i,0}} + \beta \sum_{i=2}^T \psi(Z_{i,2} - Z_{i-1,1}) \nonumber \\
&+(\rho/2)\sum_{i=1}^T \bigg(\fronorm{\Theta_i - Z_{i,0}+U_{i,0}}^2 - \fronorm{U_{i,0}}^2\bigg) \nonumber \\
&+(\rho/2)\sum_{i=2}^T \bigg( \fronorm{\Theta_{i-1} - Z_{i-1,1} + U_{i-1,1}}^2 - \fronorm{U_{i-1,1}}^2 \nonumber  \\
& \qquad \qquad+\fronorm{\Theta_{i} - Z_{i,2}+ U_{i,2}}^2 - \fronorm{U_{i,2}}^2\bigg),  \label{lagrangian} 
\end{align}
where $U = \{U_0, U_1, U_2\} = $ $\{(U_{1,0},\ldots,U_{T,0}),$ $ (U_{1,1},\ldots,U_{T-1,1}),$ $ (U_{2,2},\ldots,U_{T,2}) \}$ is the scaled dual variable and $\rho > 0$ is the ADMM penalty parameter \cite[\S 3.1.1]{BPCPE:11}.
ADMM consists of the following updates, where $k$ denotes the iteration number: 
\begin{align*}
\small
(a) &\quad \Theta^{k+1} := \argmin{\Theta \in \symmpd{p}}\mathcal{L}_\rho\left(\Theta, Z^{k}, U^{k}\right)\\
(b) &\quad Z^{k+1} = \bmat{Z_0^{k+1} \\ Z_1^{k+1} \\ Z_2^{k+1} } := \argmin{Z_0,Z_1,Z_2}\mathcal{L}_\rho\left(\Theta^{k+1}, Z, U^{k}\right)\\
(c) &\quad U^{k+1} = \bmat{U_0^{k+1} \\ U_1^{k+1} \\ U_2^{k+1}  }:=\bmat{U_0^{k}\\ U_1^{k} \\ U_2^{k} } + \bmat{ \Theta^{k+1} - Z_0^{k+1}\\  (\Theta_{1}^{k+1},\ldots,\Theta_{T-1}^{k+1}) - Z_1^{k+1} \\ (\Theta_{2}^{k+1},\ldots,\Theta_{T}^{k+1}) - Z_2^{k+1} }.
\end{align*}

\xhdr{Global Convergence}
By separating Problem \eqref{original} into two blocks of variables, $\Theta$ and $Z$, our ADMM approach is guaranteed to converge to the global optimum. 
Our iterative algorithm uses a stopping criterion based on the primal and dual residual values being below specified thresholds; see \cite{BPCPE:11}. 

\subsection{$\Theta$-Update} 
The $\Theta$-step can be split into separate updates for each $\Theta_i$, which can then be solved in parallel:
\begin{align*}
\small
&\Theta_i^{k+1} \overset{(a)}{=} \argmin{\Theta_i \in \symmpd{p}} - \log\det(\Theta_i) + \Tr{S_i \Theta_i} + \frac{1}{2\eta} \fronorm{\Theta_i - A}^2 \\
&\overset{(b)}{=} \argmin{\Theta_i \in \symmpd{p}} - \log\det(\Theta_i) + \Tr{S_i \Theta_i} +  \frac{1}{2\eta} \fronorm{\Theta_i -\frac{A + A^T}{2}}^2
\end{align*}
where (a) holds for $A =\frac{Z_{i,0}^k +Z_{i,1}^k +Z_{i,2}^k - U_{i,0}^k-U_{i,1}^k-U_{i,2}^k}{3}$ and $\eta = \frac{n_i}{3 \rho}$,
and (b) holds due to the symmetry of $\Theta_i$. 
This can be rewritten as a proximal operator, 
\begin{align*}
\small
& \Theta_i^{k+1} = \prox_{\eta (- \log\det(\cdot) + \Tr{S_i \cdot})}\left(({A+ A^T})/{2}\right).
\end{align*}
Since $\frac{A+ A^T}{2}$ is symmetric, this has an analytical solution, 
\begin{align}
\small
\Theta_i^{k+1} &:= \frac{1}{2\eta^{-1}}Q\big(D+\sqrt{D^2 + 4\eta^{-1} I}\big)Q^T,
\end{align}
where $QDQ^T$ is the eigendecomposition of $\eta^{-1} \frac{A+A^T}{2}-S_i$~\cite{DWW:14, WT:09}. This is the most computationally expensive task in our algorithm, as decomposing a $p\times p$ matrix has an $O(p^3)$ runtime.

\subsection{$Z$-Update}
The $Z$-update can be split into two parts: $Z_0$, which refers to the $\|\Theta\|_{\mathrm{od},1}$-penalty that enforces sparsity in the inverse covariance matrices, and $(Z_1,Z_2)$, which denotes the $\psi$-penalty that minimizes deviations across timestamps. These two updates can be solved simultaneously, and each part can be parallelized even further to speed up computation.

\subsubsection{Part 1: $Z_0$-Update}
Each $Z_{i,0}$ can be written as the proximal operator of the $\ell_{\mathrm{od},1}$-norm, which has a known closed-form solution \cite{PB:14}
\begin{align*}
\small
Z_{i,0}^{k+1} &= \prox_{ \frac{\lambda}{\rho}\odnorm{\cdot}}(\Theta_i^{k+1}+U_{i,0}^k) = S_{\frac{\lambda}{\rho}}(\Theta_i^{k+1}+U_{i,0}^k),
\end{align*}
where   $S_{\frac{\lambda}{\rho}}(\cdot)$ is the element-wise soft-threshold function, for $1\leq i \neq j \leq p$. The $(i,j)$-th element of this update is
\begin{align*}
\small
\left(S_{\frac{\lambda}{\rho}}(A)\right)_{ij} = 
\begin{cases}
0 \quad & |A_{ij}| \leq \frac{\lambda}{\rho} \\
\mathbf{sgn}(A_{ij})(|A_{ij}|-\frac{\lambda}{\rho}) \quad &\mbox{otherwise}. \\
\end{cases}
\end{align*}

\subsubsection{Part 2: $(Z_1$, $Z_2)$-Update}
$Z_{i-1, 1}$ and $Z_{i,2}$ are coupled together in the augmented Lagrangian, so they must be jointly updated. In order to derive the closed-form solution, we define
\begin{align*}
\small
\tilde \psi(\bmat{Z_1 \\ Z_2}) = \psi(Z_2 - Z_1).
\end{align*}
We now solve a separate update for each $(Z_1$, $Z_2)$ pair,
\begin{align}
\small
\bmat{ Z_{i-1,1}^{k+1} \\ Z_{i,2}^{k+1}} &= \prox_{\frac{\beta}{\rho} \tilde \psi(\cdot)}\left( \bmat{\Theta_{i-1}^{k+1}+U_{i-1,1}^k\\  \Theta_{i}^{k+1} + U_{i,2}^k} \right).
\label{ADMM_z12_proximal}
\end{align}

\paragraph{Part 2-1: $(Z_1,Z_2)$-Updates for the Sum of Column Norms}
For the $\ell_1, \ell_2, \ell_2^2$, and  $\ell_\infty$ penalties defined in Section 2.1, we solve the proximal operator by utilizing the following two properties~\cite{PB:14,LV:10}:
\begin{itemize}[nolistsep, leftmargin=.15in]
\vspace{1mm}
\item If a function $f$ is a composition of another function $g$ with an orthogonal affine transformation, i.e., $f(x)= g(Cx+D)$ and $CC^T= (1/\alpha)I$, then 
\begin{align*}
	\small
	\prox_f(x) = (I-\alpha C^TC)x + \alpha C^T(\prox_{\frac{1}{\alpha}g}(Cx+D)-D).
\end{align*}
\item If $f$ is block-separable, i.e., $f(x)= \sum_{j} f_j(x_j)$ where $x = (x_1, x_2, \ldots)$, then
$\left( \prox_f(v) \right)_j = \prox_{f_j}(v_j)$.
\end{itemize}
\vspace{1mm}

Recall that our goal is to get the analytical solution to (\ref{ADMM_z12_proximal}). To do so, we apply the first property  
with $f = \tilde \psi$, $g = \psi$, $C = \bmat{-I & I}$, $D = 0$, and $\alpha = \frac12$, which converts the $(Z_1, Z_2)$-update into 
\begin{align*}
\small
\bmat{ Z_{i-1,1}^{k+1} \\ Z_{i,2}^{k+1}}
&\overset{(a)}= \frac{1}{2}\bmat{\Theta_{i-1}^{k+1} + \Theta_{i}^{k+1} +U_{i-1,1}^k + U_{i,2}^k \\ \Theta_{i-1}^{k+1} + \Theta_{i}^{k+1} +U_{i-1,1}^k + U_{i,2}^k} + \frac{1}{2}\bmat{-E \\ E},
\end{align*}
where (a) holds for  
\begin{align*}
\small
E = \prox_{\frac{2\beta}{\rho} \psi}(\bmat{\Theta_{i}^{k+1} - \Theta_{i-1}^{k+1}  + U_{i,2}^k -  U_{i-1,1}^k }).
\end{align*}

Now, for each penalty function $\psi$, we simply need to solve for the corresponding $E$. We denote $A = \bmat{\Theta_{i-1}^{k+1} - \Theta_{i}^{k+1} +U_{i-1,1}^k- U_{i,2}^k}$ and  $\eta = \frac{2\beta}{\rho}$. Then, since $\psi$ is just the sum of column-norms and thus block-separable, we simplify $E$ as
\begin{align}
\small
[E]_j = \left(\prox_{\eta \psi}(A)\right)_j \overset{(b)}{=} \prox_{\eta \phi}([A]_j), \label{E_j}
\end{align}
where $\phi$ is the column-norm for $\ell_1$, $\ell_2$, $\ell_2^2$, and $\ell_{\infty}$.
Note that we have narrowed the $(Z_{i-1,1}, Z_{i,2})$-update in (\ref{ADMM_z12_proximal}) down to just finding $E_j$ in (\ref{E_j}), 
expressed by the proximal operator of $\phi$ defined on a vector, each of which has a closed form solution as follows:

\xhdr{Element-wise $\ell_1$ Penalty}
The $\ell_1$ proximal operator is
\begin{align*}
\small
[E]_{j} = \prox_{\eta \|\cdot\|_1}([A]_j)  = S_{\eta}(A_{j}).
\end{align*}
This is just the element-wise soft threshold,
\begin{align*}
\small
E_{ij} = S_{\eta}(A_{ij}) = 
\begin{cases}
0 \quad & |A_{ij}| \leq \eta \\
\mathbf{sgn}(A_{ij})(|A_{ij}|-\eta) \quad &\mbox{otherwise}.
\end{cases}
\end{align*}

\xhdr{Group Lasso $\ell_2$ Penalty}
The proximal operator for the $\ell_2$-norm is a block-wise soft thresholding,
\begin{align*}
\small
[E]_j = \prox_{\eta \|\cdot\|_2}([A]_j) &= 
\begin{cases}
0 \quad & \|[A]_j\|_2 \leq \eta \\
(1-{\eta}/{\|[A]_j\|_2})[A]_j \quad &\mbox{otherwise}.
\end{cases} \label{prox_twonorm}
\end{align*}

\xhdr{Laplacian Penalty}
The proximal operator for the $\ell_2^2$-norm, Laplacian regularization, is given by 
\begin{align*}
\small
[E]_{j} = \prox_{\eta \|\cdot\|_2^2}([A]_j) &= 
(1+2\eta)^{-1}([A]_{j}).
\end{align*}
This can be rewritten in element-wise form 
\begin{align*}
\small
E_{ij} = (1+2\eta)^{-1}(A_{ij}).
\end{align*}

\xhdr{$\ell_{\infty}$ Penalty}
The proximal operator for the $\ell_\infty$-norm is
\begin{align*}
\small
[E]_j = \prox_{\eta \|\cdot\|_\infty}([A]_j)= 
\begin{cases}
0 \quad &\|[A]_j\|_1 \leq \eta\\
[A]_j - \eta S_{\sigma}([A]_j/\eta) \quad &\mbox{ otherwise},
\end{cases}
\end{align*}
where $\sigma$ is the solution to $\sum_{i=1}^n\max\{A_{ij}/\eta- \sigma,0 \} =  1$, which has no closed-form solution but can be solved via bisection.

\paragraph{Part 2-2: $(Z_1, Z_2)$-Update for Perturbed Node Penalty}

The perturbed node proximal operator does not have an efficient analytical solution. However, we can solve this new problem by deriving a second ADMM algorithm. Here, in each iteration of our original ADMM solution, we now call this second ADMM solver. 

In order to avoid notational conflict, we denote the minimization variables $(Z_{i-1,1}, Z_{i,2})$ as $(Y_1, Y_2)$. We then introduce an additional variable $V =  W^T$, and the augmented Lagrangian $\mathcal{L}_\rho$ becomes
\begin{align*}
\small
\mathcal{L}_\rho&(V,W,Y_1,Y_2) = \beta\twonorm{V} +  \frac{\rho}{2} \fronorm{\bmat{Y_{1} \\ Y_{2}} - \bmat{\Theta_{i-1}^{k+1}+U_{i-1,1}^k\\  \Theta_{i}^{k+1} + U_{i,2}^k}}^2 \\
&+\frac{\rho}{2} \fronorm{V + W - (Y_{1} - Y_{2}) + \tilde U_1}^2 +  \frac{\rho}{2} \fronorm{V -  W^T + \tilde U_2}^2,
\end{align*} 
where $(\tilde U_1, \tilde U_2)$ is the scaled dual variable and $\rho$ is the same ADMM penalty parameter as outer ADMM.
At the $l$-th iteration, the three steps in the ADMM update are as follows:
\begin{align*}
\small
(a) \quad V&^{l+1}= \prox_{\frac{\beta}{2\rho} \twonorm{\cdot}}\big(\frac{Y_{1}^{l} - Y_{2}^{l} - W^l - \tilde U_1^l + ((W^l)^T -  \tilde U_2^l)^T}{2}\big),
\end{align*}
which has the following closed form solution for the $j$th column, with $A = \frac{Y_{1}^{l} - Y_{2}^{l} - W^l -\tilde U_1^l + ((W^l)^T -  \tilde U_2^l)^T}{2}$,
\begin{align*}
\small
[V^{l+1}]_j = 
\begin{cases}
0 & \|[A]_j\|_2 \leq \frac{\beta}{2\rho} \\
(1-1/({2\rho\|[A]_j\|_2}))[A]_j \quad  \quad & \mbox{otherwise}.
\end{cases}
\end{align*}
\begin{align*}
\small
(b)  \bmat{W^{l+1} \\ Y_1^{l+1} \\ Y_2^{l+1}} =  (C^TC + 2 I)^{-1} \left(2 \bmat{(V^l + \tilde U_2^l)^T \\ \Theta_{i-1}^{k+1}+U_{i-1,1}^k\\  \Theta_{i}^{k+1} + U_{i,2}^k} - C^TD\right),
\end{align*}
where $C = \bmat{I & -I & I}$, and $D =(V^l + \tilde U_1^l) $. 
\begin{align*}
\small
(c) & \bmat{\tilde U_1^{l+1} \\ \tilde U_2^{l+1}} = \bmat{\tilde U_1^{l} \\ \tilde U_2^{l}}+ \bmat{ (V^{l+1} + W^{l+1}) - (Y_1^{l+1} -  Y_2^{l+1}))\\ V^{l+1} - (W^{l+1})^T }.
\end{align*}

\section{Implementation}
\label{sec:implementation}
% !TEX root = paper-netinf.tex

We have built a custom TVGL Python solver\footnote{Code and solver can be found at http://snap.stanford.edu/tvgl/.} on top of SnapVX \cite{snapvx}, an open-source convex optimization package. Our solver takes as inputs the multivariate observations, the regularization parameters, and the type of penalty ($\ell_1$, $\ell_2$, Laplacian, $\ell_{\infty}$, or perturbed node), and it returns the time-varying network. Although TVGL is capable of being distributed across many machines, we instead distribute it across multiple cores of a single large-memory machine.

\section{Experiments}
\label{sec:experiments}
% !TEX root = paper-netinf.tex

Here, we run several experiments on synthetic data, where there are clear ground truth networks, to test the accuracy and scalability of our network inference approach. First, we compare our TVGL method to two state-of-the-art baselines to measure accuracy, and we demonstrate the importance of using appropriate penalties for different types of temporal evolutions. Next, we vary the problem size over several orders of magnitude and test our ADMM-based algorithm's scalability compared to three other solution methods.

\subsection{Accuracy on Synthetic Data}
We first analyze a synthetic problem in which the observations are generated from a changing underlying covariance matrix. This provides a known ground truth network, which we can use to verify the accuracy of our network inference approach.

\xhdr{Experimental Setup}
We evaluate two different types of temporal evolutions: a global shift, where the entire structure of the network changes at some time $t$, and a single node perturbation (which we refer to as a local shift), where one node rewires its connections all at once but the rest of the graph remains the same. We randomly generate the ground truth covariance and subsequent samples using the method outlined by Mohan et al.~\cite{MLFWL:14}.
For both examples, we generate data in $\reals^{10}$ over $100$ timestamps, where the shift (either global or local) occurs at time $t = 50$. At each $t$, we observe 10 independent samples from the true distribution. 

From this time series of observations, we then solve for $\Theta_t, t = 1,\ldots,100$, our estimate of the dynamic network across this time period.
Here, we set the regularization parameters $\lambda$ and $\beta$ as the values that minimize the Akaike Information Criteria (AIC)~\cite{HTF:09} (on a separate, independently generated training set).

\xhdr{Baseline Methods}
We compare our approach to two different baselines: the static graphical lasso~\cite{FHT:08} and the kernel method~\cite{zhou2010time}.
For the static graphical lasso, we treat each $t_i$ as an independent network. Since our data has 100 time steps, this means we solve 100 independent graphical lasso problems and infer 100 separate networks. For the kernel method, we modify the empirical covariances and weight them according to a non-negative kernel function. We set the kernel width to the theoretically guaranteed optimum using the method proposed by Zhou et al.~\cite{zhou2010time}.

\xhdr{Performance Measures}
We introduce two metrics to measure the accuracy of our estimate:
\begin{itemize}[nolistsep, leftmargin=.15in]
    \item \textbf{$F_1$ score:} This measures how closely we capture the true edge structure of the network (\ie, how many of the non-zero elements in the inverse covariance we correctly identify as non-zero). This score is the harmonic mean of the precision and recall. 
    \item \textbf{Temporal deviation (TD) ratio:} The temporal deviation, $\|\Theta_i -  \Theta_{i-1}\|_F$, shows how much the estimate has changed at each timestamp. This score is the ratio of the temporal deviation at $t=50$ (where the one ``true'' shift happened) to the average temporal deviation value across the 100 timestamps.
\end{itemize}

\xhdr{Experimental Results}
We show results for both the local and global shift with several different TVGL penalties in Table \ref{synTable}. In terms of both $F_1$ score and temporal deviation, our TVGL approach significantly outperforms the two baselines.
The TVGL $F_1$ score is up to 38.4\% higher than the kernel method and 91.9\% higher than the static graphical lasso. Regardless of penalty type, the TVGL also always has a temporal deviation (TD) ratio at least 9.7 times larger than any of the baseline methods. In fact, for both baselines and both shift types, the largest temporal deviation peak in the time series does \emph{not} occur at $t=50$. This means that the baselines do not detect that there is a large shift in the network at this time, whereas this sudden change is clearly discovered by TVGL (where the largest peak always occurs at $t=50$). We later use this idea in Section 7 to detect significant events in real-world time series data.

\begin{table}[t]
\centering
\resizebox{8.6cm}{!} {
    \begin{tabular}{c | c | c | c | c | c | c }
    \hline
    True & Score & Static GL & Kernel & TVGL & TVGL & TVGL \\ 
    Shift & & & & ($\ell_1$) & ($\ell_2$) & (Perturbed Node)
    \\\hline
    \multirow{2}{*}{Local} & $F_1$ & 0.646 & 0.768 & 0.819 & 0.817 & \textbf{0.853} \\ \cline{3-7}
    \multirow{2}{*}{} & TD ratio & 2.02 & 2.41 & 27.9 & 23.3 & \textbf{55.5} \\
     \hline
     \multirow{2}{*}{Global} & $F_1$ & 0.496 & 0.688 & 0.939 & \textbf{0.952} & 0.943 \\ \cline{3-7}
    \multirow{2}{*}{} & TD ratio & 1.06 & 1.80 & \textbf{47.6} & 38.6 & 36.2 \\ \hline
    \end{tabular}
    }
    \vspace{2mm}
    \caption{$F_1$ score and Temporal Deviation (TD) ratio for a local and global shift, using the two baselines and three different TVGL evolutionary penalties.}
    \label{synTable}
  \vspace{-11mm}
\end{table}

\xhdr{Selection of Penalty Type}
While the TVGL outperformed the two baselines regardless of the penalty type, even greater gains can be achieved by selecting the correct evolutionary penalty. In real world cases, this parameter can be selected by cross-validation or by incorporating domain knowledge, using the descriptions in Section 2.1 to choose the proper penalty based on exactly what type of temporal evolution one is looking for in the data. As shown in Table \ref{synTable}, there are clear benefits from using certain penalties in certain situations. For example, with a local shift, which is well-suited to be analyzed by a perturbed node penalty, choosing this penalty leads to a 5\% higher $F_1$ score and a temporal deviation ratio that is more than twice as large as both the $\ell_1$ or $\ell_2$ cases.
For the Global shift, the $\ell_2$ penalty does the best job at reconstructing the time-varying network (largest $F_1$ score), though the $\ell_1$ is better able to identify the sudden shift at $t=50$ (with its TD ratio 23\% larger than either of the other two penalty types).
This additional selection parameter, which to the best of our knowledge has not been previously explored in this context, expands the reach of time-varying inference methods by allowing us to model various types of network evolutions with high precision.

\subsection{Scalability of TVGL}

Next, we examine the scalability of our TVGL algorithm. 
We run our TVGL solver on data generated by the method outlined in Section 6.1, and we vary our problem size over several orders of magnitude. Here, we estimate $m$ slices of a $n$-node network, for a total of $m\frac{n(n+1)}{2}$ unknown variables.

While many algorithms exist to efficiently solve the static graphical lasso problem (\eg, \cite{FHT:08, hsieh2013big}), these cannot be directly applied to our problem because of the time-varying penalties coupling the variables together. Instead, we compare our runtime against 
three alternative methods that can solve our time-varying problem: two semidefinite programming solvers (CVXOPT~\cite{DV:06} and SCS~\cite{OCPB:16}) and a naive ADMM method (without the closed-form updates that we developed in \S4).
We experiment on a single 40-core CPU where the entire problem fits into memory. We set $m = 10$ and modify $n$ to vary the problem size. Even though our TVGL algorithm can solve for problems with much larger values of $m$, it is intractable to scale the other methods beyond this point, and we run the experiments in identical conditions to isolate our algorithm's effect on scalability.

\begin{figure}
    \centering
    \includegraphics[width=0.97\linewidth]{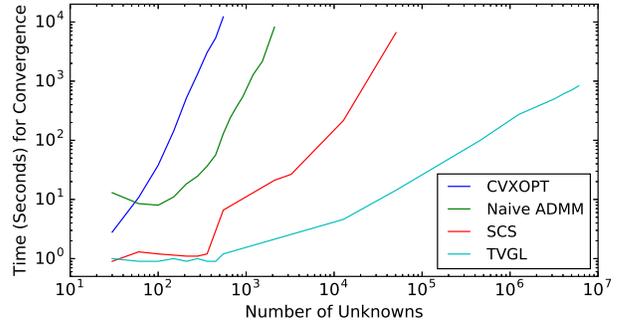}
    \vspace{-4mm}
    \caption{Scalability comparison between our TVGL solver and three other solution methods.}
    \vspace{-3mm}
    \label{fig:scalability}
\end{figure}

We compare the performance of the four solvers in Figure \ref{fig:scalability}. 
As shown, problems which take hours for the other solvers can be solved in seconds using TVGL. For example, to solve for 50,000 unknowns, TVGL is over 400 times faster than SCS, the second fastest solver (14.3 vs.\ 6,611 seconds).
The main difference in solution time is due to the closed-form ADMM solutions we derived in Section 4.1 for each of the ADMM subproblems. Our problem has a semidefinite programming (SDP) constraint, which is particularly difficult to solve via standard methods. To infer a single $n$-by-$n$ covariance, the largest per-iteration cost in our algorithm is $O(n^3)$, the cost of an eigendecomposition during the $\Theta$-update. For the same problem, general interior-point methods have a runtime of $O(n^6)$~\cite{MLFWL:14}. 
These numbers empirically appear to hold true in Figure \ref{fig:scalability} (recall that the total number of unknowns in the x-axis scales with $O(n^2)$).

\section{Case Studies}
\label{sec:applications}
% !TEX root = paper-netinf.tex

We next apply the TVGL to two real-world domains to illustrate several basic examples of how our approach can be used to learn meaningful insights from multivariate time series data.

\subsection{Applications in Financial Data}

By examining historical stock prices, we can infer a financial network to model relationships between different companies. Learning the structure of this network is useful because it allows us to devise models of how certain stocks are related, which can be used to predict future prices, understand economic trends, or even diversify a portfolio by avoiding highly correlated stocks. We infer this network by examining the stock prices of several large companies in 2010. 
We apply our method by treating the closing price of each stock as a daily ``sensor'' observation. 
Our inferred graphical representation then shows how the stock prices affect each other. 

\begin{figure}
    \centering
    \includegraphics[width=0.99\linewidth]{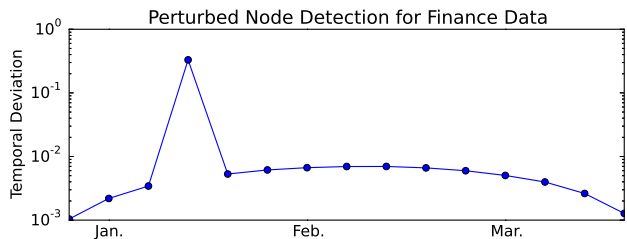}
    \vspace{-5mm}
    \caption{Plot of the temporal deviation of our estimated stock network, which detects a local shift in late January. 
    }
    \vspace{-3mm}
    \label{fig:applePlot}
    \vspace{-5mm}
\end{figure}

\begin{figure}[!t]
\centering 
  \subfigure[]{\label{fig:stock1}\centering\includegraphics[width=0.44\linewidth]{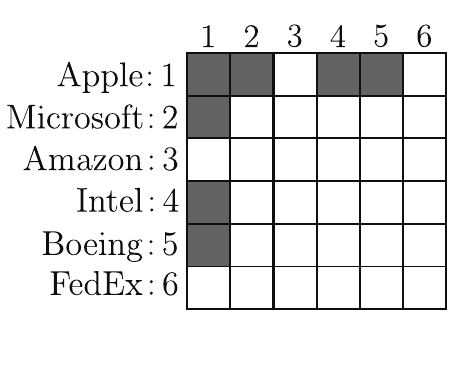}}
  \subfigure[]{\label{fig:stock2}\centering\includegraphics[width=0.53\linewidth]{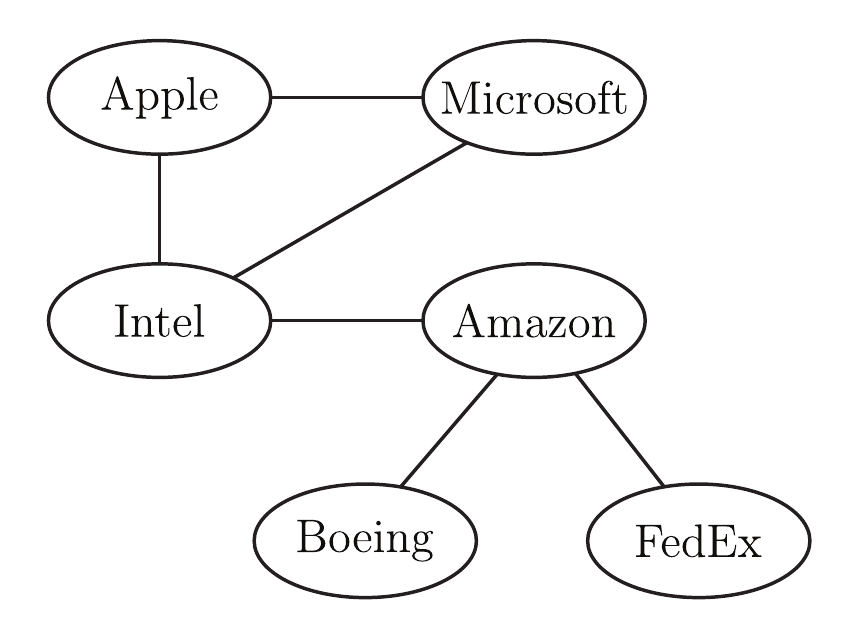}}
  \vspace{-5mm}
    \caption{Sparsity structure of (a) the network change during the local shift, indicating that Apple is the perturbed node, (b) the relationships in the financial network.}
   \vspace{-3mm}
   \label{fig:stocks}
   \vspace{-1mm}
\end{figure}

\xhdr{Identifying Single-Stock Perturbations}
We observe daily stock prices
for six large American companies: Apple, Google, Amazon, Intel, Boeing, and FedEx. 
Generally, large companies are well-established in the market, so we would expect their correlation network to change only slightly over time. However, events occasionally occur which cause a sudden shift in the network structure. We look at the dependencies of these companies while using a perturbed node penalty. This enforces a temporal dynamic where single nodes may occasionally reweigh all their connections at once. This typically reflects that something happened to affect just one company, while leaving the rest of the network unchanged. 
We solve the TVGL optimization problem with a perturbed node penalty and discover that one event stood out as having the \emph{largest single-node effect} on the dynamics between these stocks.

After running our TVGL method, we plot the temporal deviation, $\|\Theta_i -  \Theta_{i-1}\|_F$, in Figure \ref{fig:applePlot}. We discover that there is a large spike in the temporal deviation score during the last week of January. This represents a significant ``shift'' in the network at this specific time.
We show the network change at this shift in Figure \ref{fig:stock1}, where we see that the perturbed stock was Apple. At this timestamp, only Apple's edges were affected, reflecting a local rather than global change in the network. 
We examined the media to understand what may have caused this shift, and we found that on January 27th, Apple first introduced the original iPad to the public. This corresponds to the exact time that our TVGL method captured a structural change in the network, where it was also able to identify Apple as the cause of this shift.

We also plot the post-announcement network in Figure \ref{fig:stock2}. Analyzing the correlation structure yields some insightful relationships. The four technology companies---Apple, Google, Amazon, and Intel---are closely related. The shipping company (FedEx) and the airplane manufacturer (Boeing) are both connected to only one company, Amazon. Amazon heavily depends on both companies to deliver products on time, while both rely on Amazon for a large portion of their business. Interestingly, our model predicts that FedEx and Boeing's stock prices are conditionally independent given Amazon, a trend that holds true across the entire dataset.

\xhdr{Detecting Large-Scale Events}
Time-varying inference can also be used to detect significant events that affected the entire network, not just single entities within it. We run a similar experiment to the previous example, except we now include every company in the S\&P 500, which covers 500 of the largest US-based public companies. 
Since we are focusing on macro-level event detection, we look for the maximum temporal deviation with an $\ell_1$ penalty. This point in time represents the largest ``shock'' to the network, the timestamp where our TVGL model detected a large and sudden change. We discover that a shift happens during the week of May 6th, 2010. Surprisingly, we saw that there was in fact a ``Flash Crash'' that day, where the entire market dropped 9\% in a matter of seconds, only to rebound back just minutes later. Our results imply that, even though the market recovered from the crash in terms of stock price values, there were tangible long-term effects on the correlation network between companies in the S\&P 500.

\subsection{Application to Automobile Sensors}
Inferring relationships between interrelated entities is of particular interest to industries with large amounts of sensor data. One such industry is automobiles, where modern cars contain hundreds of sensors measuring everything from the car's velocity to the slope of the road. Inferring a time-varying network of relationships between these sensors is an informative way to understand driving habits, as each driver creates a unique ``signature'', or dynamic sensor network, while driving. This network can be used to model behavior, compare driver ability, or even detect impaired drivers.

\begin{figure}[!t]
    \centering
    \includegraphics[width=0.99\linewidth]{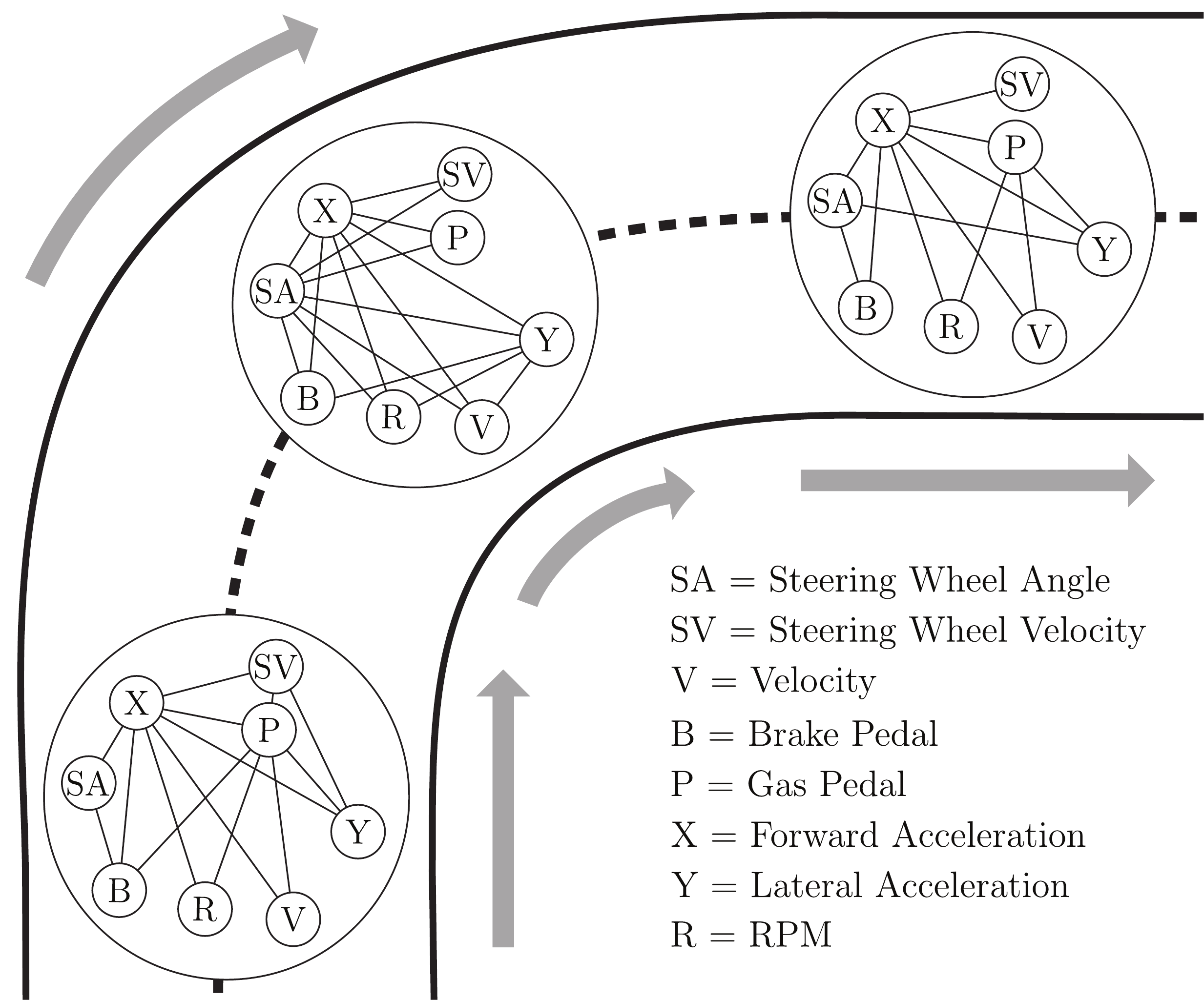}
    \vspace{-4mm}
    \caption{Three snapshots of the automobile sensor network measuring eight sensors, taken (1) before, (2) during, and (3) after a standard right turn.}
    \vspace{-5mm}
    \label{fig:cars}
\end{figure}

As a second case study, we analyze an automobile sensor dataset, provided by a large car company. 
We look at a short session where a driver goes down a straight road, makes a right turn, and continues on a new road. 
We observe eight different sensors every 0.1 seconds: steering wheel angle, steering wheel velocity, vehicle velocity, brake pedal, gas pedal, forward acceleration, lateral acceleration, and engine RPM.
Since we do not expect any sudden shifts in this network, we use a Laplacian penalty to enforce a smoothly varying network across time. We run our TVGL algorithm and plot three snapshots of the sensor network in Figure \ref{fig:cars}: one on the straightaway before the turn, one in the middle of turning, and one after the turn. Note that the ``before'' and ``after'' networks look quite similar, since the driver is doing a similar activity (driving straight) on both. The ``during'' network looks much different, however. The most noticeable difference is that the steering wheel angle sensor (SA) is now centrally located within the network, connected to almost every other node. This means that variations in the other sensors' observations can be largely explained by the steering wheel angle during the turn. For straightaways, on the other hand, the steering angle is on the network periphery, meaning that it is not well suited to explain the other sensor readings.

\section{Conclusion and Future Work}
\label{sec:conclusion}
% !TEX root = paper-netinf.tex

In this paper, we have defined a general method for inferring dynamic networks from timestamped observational data. Our approach, the time-varying graphical lasso, provides a scalable algorithm capable of encoding many different temporal dependencies. This type of modeling allows us to infer structure from large-scale ``sensor'' deployments. 
We leave for future work the examination of additional penalty functions to enforce different behaviors in the evolution of the network structure, along with their closed-form proximal operator solutions. There are also new extensions, beyond what was discussed in Section 3, which could be further analyzed. For example, we currently assume that the underlying distribution is zero mean. However, we could model a problem where observations come from a distribution $x \thicksim \mathcal{N}(\mu(t),\Sigma(t))$, and attempt to simultaneously estimate both the mean and covariance as they vary over time. 
Finally, this work could also be extended to infer correlations across different timestamps. Currently, our algorithm is suited to find simultaneously-related entities, \ie, stocks whose prices move up and down in unison. However, there are other applications where the correlations are not immediate. For example, in brain connectivity networks, neuron $A$ firing at timestamp $i$ could cause neuron $B$ to fire at time $i+1$. Each of these additions would open up our framework to new potential applications, providing additional benefits to future research in this topic. 

\xhdr{Acknowledgements} This work was supported by NSF IIS-1149837, NIH BD2K, DARPA SIMPLEX, DARPA XDATA, Chan Zuckerberg Biohub, SDSI, Boeing, Bosch, and Volkswagen.

\bibliography{refs}

\begin{thebibliography}{10}

\bibitem{AX:09}
A.~Ahmed and E.~P. Xing.
\newblock Recovering time-varying networks of dependencies in social and
  biological studies.
\newblock {\em PNAS}, 2009.

\bibitem{BED:08}
O.~Banerjee, L.~El~Ghaoui, and A.~d'Aspremont.
\newblock Model selection through sparse maximum likelihood estimation for
  multivariate {Gaussian} or binary data.
\newblock {\em JMLR}, 2008.

\bibitem{BPCPE:11}
S.~Boyd, N.~Parikh, E.~Chu, B.~Peleato, and J.~Eckstein.
\newblock Distributed optimization and statistical learning via the alternating
  direction method of multipliers.
\newblock {\em Foundations and Trends in Machine learning}, 2011.

\bibitem{BV:04}
S.~Boyd and L.~Vandenberghe.
\newblock {\em Convex Optimization}.
\newblock Cambridge University Press, 2004.

\bibitem{DV:06}
J.~Dahl and L.~Vandenberghe.
\newblock {CVXOPT}: A {P}ython package for convex optimization.
\newblock In {\em Proc. Eur. Conf. Op. Res}, 2006.

\bibitem{DWW:14}
P.~Danaher, P.~Wang, and D.~Witten.
\newblock The joint graphical lasso for inverse covariance estimation across
  multiple classes.
\newblock {\em JRSS: Series B}, 2014.

\bibitem{FHT:08}
J.~Friedman, T.~Hastie, and R.~Tibshirani.
\newblock Sparse inverse covariance estimation with the graphical lasso.
\newblock {\em Biostatistics}, 2008.

\bibitem{gomez12netinf}
M.~Gomez~Rodriguez, J.~Leskovec, and A.~Krause.
\newblock Inferring networks of diffusion and influence.
\newblock In {\em KDD}. ACM, 2010.

\bibitem{grewal2011kalman}
M.~S. Grewal.
\newblock {\em Kalman filtering}.
\newblock Springer, 2011.

\bibitem{HLB:15}
D.~Hallac, J.~Leskovec, and S.~Boyd.
\newblock Network lasso: Clustering and optimization in large graphs.
\newblock In {\em KDD}, 2015.

\bibitem{snapvx}
D.~Hallac, C.~Wong, S.~Diamond, R.~Sosi\v{c}, S.~Boyd, and J.~Leskovec.
\newblock {SnapVX}: A network-based convex optimization solver.
\newblock {\em JMLR (To Appear)}, 2017.

\bibitem{HTF:09}
T.~Hastie, R.~Tibshirani, and J.~Friedman.
\newblock {\em The Elements of Statistical Learning}.
\newblock Springer, 2009.

\bibitem{H:69}
M.~R. Hestenes.
\newblock Multiplier and gradient methods.
\newblock {\em Journal of Optimization Theory and Applications}, 1969.

\bibitem{hsieh2013big}
C.-J. Hsieh, M.~A. Sustik, I.~S. Dhillon, P.~K. Ravikumar, and R.~Poldrack.
\newblock {BIG} \& {QUIC}: Sparse inverse covariance estimation for a million
  variables.
\newblock In {\em NIPS}, 2013.

\bibitem{kolar2010estimating}
M.~Kolar, L.~Song, A.~Ahmed, and E.~Xing.
\newblock Estimating time-varying networks.
\newblock {\em The Annals of Applied Statistics}, 2010.

\bibitem{KF:09}
D.~Koller and N.~Friedman.
\newblock {\em Probabilistic Graphical Models: Principles and Techniques}.
\newblock MIT press, 2009.

\bibitem{L:96}
S.~L. Lauritzen.
\newblock {\em Graphical models}.
\newblock Clarendon Press, 1996.

\bibitem{MCHWLF:12}
K.~Mohan, M.~Chung, S.~Han, D.~Witten, S.-I. Lee, and M.~Fazel.
\newblock Structured learning of {G}aussian graphical models.
\newblock In {\em NIPS}, 2012.

\bibitem{MLFWL:14}
K.~Mohan, P.~London, M.~Fazel, D.~Witten, and S.-I. Lee.
\newblock Node-based learning of multiple {G}aussian graphical models.
\newblock {\em JMLR}, 2014.

\bibitem{molenaar1985dynamic}
P.~C. Molenaar.
\newblock A dynamic factor model for the analysis of multivariate time series.
\newblock {\em Psychometrika}, 1985.

\bibitem{MHSLAM:14}
R.~P. Monti, P.~Hellyer, D.~Sharp, R.~Leech, C.~Anagnostopoulos, and
  G.~Montana.
\newblock Estimating time-varying brain connectivity networks from functional
  {MRI} time series.
\newblock {\em Neuroimage}, 2014.

\bibitem{myers10connie}
S.~Myers and J.~Leskovec.
\newblock On the convexity of latent social network inference.
\newblock In {\em NIPS}, 2010.

\bibitem{N:11}
A.~Namaki, A.~Shirazi, R.~Raei, and G.~Jafari.
\newblock Network analysis of a financial market based on genuine correlation
  and threshold method.
\newblock {\em Physica A: Stat. Mech. Apps.}, 2011.

\bibitem{OCPB:16}
B.~O'Donoghue, E.~Chu, N.~Parikh, and S.~Boyd.
\newblock Conic optimization via operator splitting and homogeneous self-dual
  embedding.
\newblock {\em Journal of Optimization Theory and Applications}, 2016.

\bibitem{PB:14}
N.~Parikh and S.~Boyd.
\newblock Proximal algorithms.
\newblock {\em Foundations and Trends in Optimization}, 2014.

\bibitem{RH:05}
H.~Rue and L.~Held.
\newblock {\em {G}aussian {M}arkov Random Fields: Theory and Applications}.
\newblock CRC Press, 2005.

\bibitem{SMG:10}
K.~Scheinberg, S.~Ma, and D.~Goldfarb.
\newblock Sparse inverse covariance selection via alternating linearization
  methods.
\newblock In {\em NIPS}, 2010.

\bibitem{LV:10}
L.~Vanderberghe.
\newblock Proximal mapping lecture notes.
\newblock \url{http://seas.ucla.edu/~vandenbe/236C/lectures/proxop.pdf}, 2010.

\bibitem{WJ:06}
M.~J. Wainwright and M.~I. Jordan.
\newblock Log-determinant relaxation for approximate inference in discrete
  {M}arkov random fields.
\newblock {\em IEEE Tr. on Signal Processing}, 2006.

\bibitem{WSZS:06}
K.~Weinberger, F.~Sha, Q.~Zhu, and L.~Saul.
\newblock Graph {L}aplacian regularization for large-scale semidefinite
  programming.
\newblock In {\em NIPS}, 2006.

\bibitem{WA:15}
E.~Wit and A.~Abbruzzo.
\newblock Inferring slowly-changing dynamic gene-regulatory networks.
\newblock {\em BMC Bioinformatics}, 2015.

\bibitem{WT:09}
D.~Witten and R.~Tibshirani.
\newblock Covariance-regularized regression and classification for high
  dimensional problems.
\newblock {\em JRSS: Series B}, 2009.

\bibitem{WK:13}
M.~Wytock and J.~Z. Kolter.
\newblock Sparse {G}aussian conditional random fields: Algorithms, theory, and
  application to energy forecasting.
\newblock {\em ICML}, 2013.

\bibitem{YLSWY:15}
S.~Yang, Z.~Lu, X.~Shen, P.~Wonka, and J.~Ye.
\newblock Fused multiple graphical lasso.
\newblock {\em SIAM}, 2015.

\bibitem{YL:07}
M.~Yuan and Y.~Lin.
\newblock Model selection and estimation in the {G}aussian graphical model.
\newblock {\em Biometrika}, 2007.

\bibitem{zhou2010time}
S.~Zhou, J.~Lafferty, and L.~Wasserman.
\newblock Time varying undirected graphs.
\newblock {\em Machine Learning}, 2010.

\end{thebibliography}
\bibliographystyle{abbrv}

\end{document}